\documentclass[letterpaper]{article} 
\usepackage{aaai24}  
\usepackage{times}  
\usepackage{helvet}  
\usepackage{courier}  
\usepackage[hyphens]{url}  
\usepackage{graphicx} 
\usepackage{hyperref}
\hypersetup{
    colorlinks=true,
    linkcolor=blue,
    filecolor=black,      
    urlcolor=blue,
    citecolor=blue
    }
\usepackage{natbib}  
\usepackage{caption} 
\usepackage{algorithm}
\usepackage{algorithmic}
\usepackage[table]{xcolor}
\usepackage{cleveref}
\crefname{figure}{Figure}{Figures}
\crefname{table}{Table}{Tables}
\crefname{section}{Section}{Sections}
\usepackage{multirow}
\usepackage{makecell}

\usepackage{newfloat}
\usepackage{listings}
\DeclareCaptionStyle{ruled}{labelfont=normalfont,labelsep=colon,strut=off} 
\floatstyle{ruled}
\newfloat{listing}{tb}{lst}{}
\floatname{listing}{Listing}
\title{D3G: Diverse Demographic Data Generation Increases Zero-Shot Image Classification Accuracy within Multimodal Models}
\author {
    Javon Hickmon
}
\affiliations{
    Department of Computer Science, University of Washington, Seattle WA\\
    javonh@cs.washington.edu
}

\usepackage{bibentry}

\begin{document}

\maketitle

\begin{abstract}
Image classification is a task essential for machine perception to achieve human-level image understanding. Multimodal models such as CLIP have been able to perform well on this task by learning semantic similarities across vision and language; however, despite these advances, image classification is still a challenging task. Models with low capacity often suffer from underfitting and thus underperform on fine-grained image classification. Along with this, it is important to ensure high-quality data with rich cross-modal representations of each class, which is often difficult to generate. When datasets do not enforce balanced demographics, the predictions will be biased toward the more represented class, while others will be neglected. We focus on how these issues can lead to harmful bias for zero-shot image classification, and explore how to combat these issues in demographic bias. We propose Diverse Demographic Data Generation (D3G), a training-free, zero-shot method of boosting classification accuracy while reducing demographic bias in pre-trained multimodal models. With this method, we utilize CLIP as our base multimodal model and Stable Diffusion XL as our generative model. We demonstrate that providing diverse demographic data at inference time improves performance for these models, and explore the impact of individual demographics on the resulting accuracy metric.
\end{abstract}

\section{Introduction}
Deep Learning systems have been a promising new paradigm in the field of image classification. Vision-Language systems are able to utilize multiple modalities to create models that are more generalizable to a broad range of downstream tasks. Despite this performance, issues such as data redundancy, noise, and class imbalance are just a few of the many difficulties that arise from collecting large amounts of training data. Multimodal models, in particular, require large amounts of high-quality training data with rich cross-modal representations in order to perform well compared to their unimodal counterparts. These models leverage the massive amounts of image-text pairs available online by learning to associate the images with their correct caption, leading to greater flexibility during inference \cite{pratt2023}; however, class imbalances can often lead to gender and racial bias depending on the desired task.\\
\\
Existing public face image datasets are strongly biased toward Caucasian faces; meanwhile, other races (i.e., Latino) are significantly underrepresented \cref{fig:face_datasets}. As a result, the models trained from such datasets suffer from inconsistent classification accuracies, limiting the applicability of such systems to predominantly white racial groups \citet{karkkainen2021fairface}. This means that minority subpopulations can potentially be further marginalized when applied to certain downstream tasks without calibration. This is a core tenet of machine learning: poor data produces poor models.\\
\begin{figure}[H]
    \begin{center}
        \includegraphics[width=7cm]{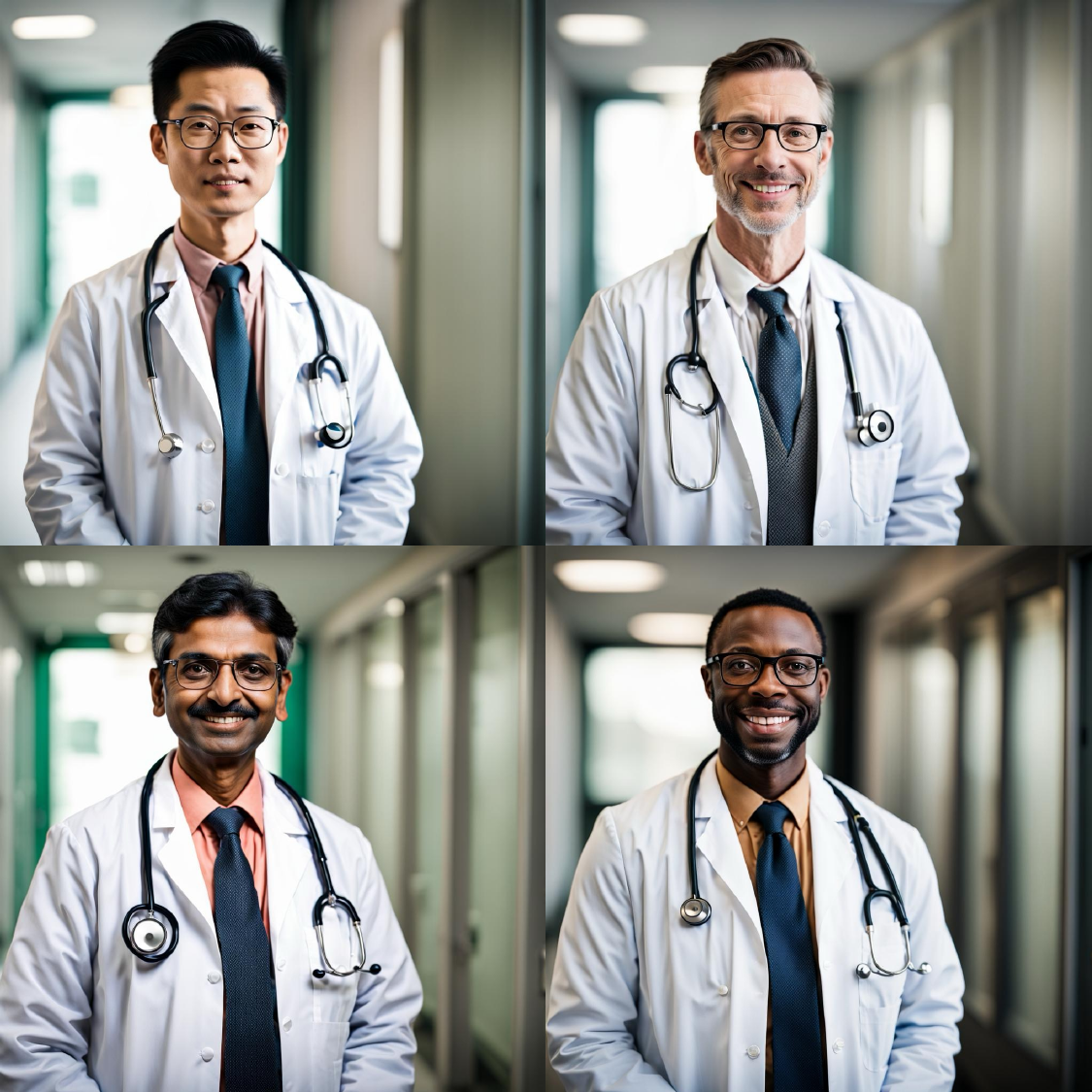}
        \caption{Images Generated with D3G for Race 4}
        \label{fig:d3g_photos}
    \end{center}
\end{figure}

\begin{figure}
    \begin{center}
        \includegraphics[width=7cm]{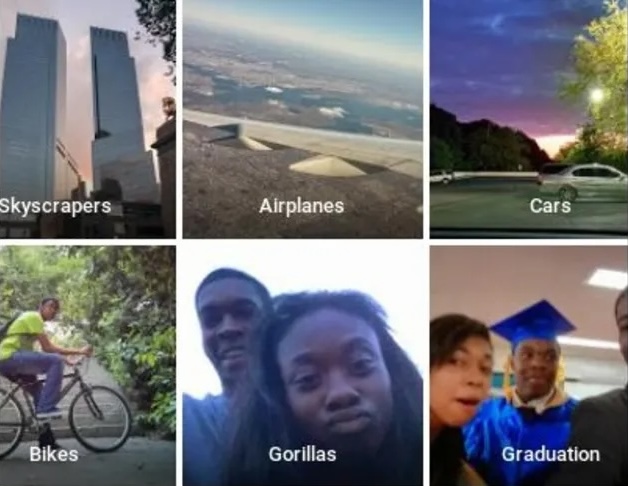}
        \caption{The image of Google Photos misclassifying black people within the Photos application.}
        \label{fig:google_photos}
    \end{center}
\end{figure}

It is important to note that we acknowledge that not all bias is harmful and often is necessary for models to generalize. This is why within this work, we focus on demographic bias which can frequently have harmful implications when applied within society. Image classification is particularly pressing, because it is the core of a myriad of computer vision tasks. Facial recognition, object detection, image search, content moderation, sentiment analysis, and many more tasks are grounded in accurate image classification systems. This is compounded by the fact that many widely used Foundational Models require multiple modalities, such as DALL-E \citet{ramesh2021zero} and Stable Diffusion \citet{rombach2022high}. In order to train these models, other models such as CLIP \citet{radford2021learning} are used to classify the data that the model will be trained on, in order to enforce a strong cross-modal correlation. This means that demographic biases will be compounded as the images continue to be utilized in training processes.\\
\\
When there are strong harmful demographic biases, these models can cause tremendous harms. One example was when Google Photo's classified black people as gorillas within a user's album, as shown in \cref{fig:google_photos}. This is the direct result of demographic bias. The dataset, used to quantify the classification accuracies of Google's model, likely contained the standard biases where images of non-Caucasian faces are underrepresented \cref{fig:face_datasets}. This leads to a misleading evaluation, which was likely why this model was deployed to the public with this significant issue.\\
\\
In this work, we explore how demographic bias affects image classification accuracy for multimodal models. We also propose D3G, a zero-shot, training-free framework to balance demographic bias and boost classification accuracies for multimodal models used for image classification.

\section{Related Work}

\subsection{Model Ensembling for Image Classification}
\begin{figure}[H]
    \begin{center}
        \includegraphics[width=8cm]{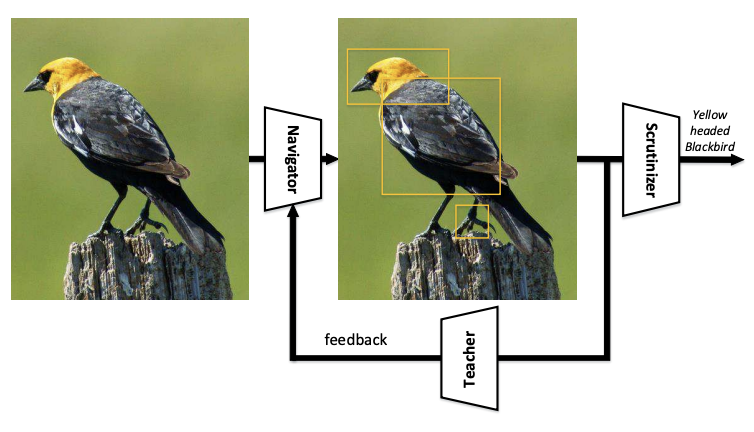}
        \caption{The NTSNet Architecture \cite{yang2018learning}}
        \label{fig:ntsnet}
    \end{center}
\end{figure}

Many state-of-the-art techniques for image classification leverage a methodology known as model ensembling. Ensemble learning broadly is an approach that aims to improve performance by combining the predictions of multiple models. There are many such ensemble learning methods, but the one most relevant to our proposed technique is called bagging.\\
\\
\textbf{\textit{Bagging predictors}} originally published by \citet{breiman1996bagging} introduced “bootstrap aggregating” or bagging, the ensemble learning method that combines multiple models trained on different subsets of the training data. The formulation is presented as follows: A learning set of $L$ consists of data ${(y_n, \mathbf{x}_n), n = 1, ..., N}$ where the $y$'s are class labels, assume we form a predictor $\phi(\mathbf{x}, L)$ where if the input is $\mathbf{x}$, we predict y by $\phi(\mathbf{x})$. Now suppose we form a sequence of replicate learning sets, ${L^{(B)}}$ each consisting of $N$ observations drawn at random with replacement from $L$. Since $y$ is a class label in our scenario, the predictions from each of the predictors ${\phi(\mathbf{x}, L^{(B)})}$ will vote to form $\phi_B(\mathbf{x})$, the aggregated final prediction. Bagging both empirically and theoretically prove improves accuracy for a given set of weak classifiers or “weak learners.” This technique effectively replicates our proposed method within a controlled environment. In the implementation of D3G, we are employing a strategy similar to bagging, but across modalities and with generated data. Our goal is based on the theoretical guarantees of bagging, where even though each model is trained on a subset of the data, the aggregation of the predictions begins to approximate the true distribution of the data.
\\
\\
A model that applies this concept of model ensembling was introduced in \textbf{\textit{Learning to Navigate for Fine-grained Classification}} by \citet{yang2018learning}. This is a state-of-the-art paper that attempts to reduce misclassification rates by developing a model called NTS-Net (Navigator-Teacher-Scrutinizer Network) to teach itself methods of identifying and scrutinizing fine-grained image details. The Navigator navigates the model to focus on the most informative regions (denoted by yellow rectangles in \cref{fig:ntsnet}), while the Teacher evaluates the regions proposed by the Navigator and provides feedback. After that, the Scrutinizer scrutinizes those regions to make predictions. NTS-Net achieves high classification accuracy on a pre-defined dataset; however, we wanted to explore if these foundational concepts could be applied to a training-free, zero-shot environment. In order to achieve this, we leverage pretrained open-vocabulary models with advanced attention mechanisms to discriminate the fine-grained features of a given image, to improve performance on a variety of classes without additional training.
Similarly, within the paper \textbf{\textit{Sus-X: Training-Free Name-Only Transfer of Vision-Language Models}} by \citet{udandarao2022sus} achieves state-of-the-art zero-shot classification results on 19 benchmark datasets, outperforming other training-free adaptation methods. It also demonstrates strong performance in the training-free few-shot setting, surpassing previous state-of-the-art methods. This paper is focused on general image classification improvements; however, we aim to explore how this idea of synthetic support set generation affects the fairness of predictions from a classification model. We will employ a similar strategy but also explore how to offset existing harmful biases within the zero-shot setting.

\subsection{Data Filtering and Generation}
\textbf{\textit{Neural Priming for Sample-Efficient Adaptation}} by \citet{wallingford2024neural} proposes a technique to adapt large pretrained models to distribution shifts. This paper demonstrates that we can leverage an open-vocabulary model's own pretraining data in order to improve performance on downstream tasks. Even though we don't aim to utilize the model's training data in our method, the generated images will likely be sampled from a similar distribution as the multimodal model. This paper shows that even if that is the case, we can still use filtering and guidance in order to improve performance. In our case, our custom prompting method plays the role of guiding the image generation process, resulting in the same empirical performance improvements.\\
\\
\textbf{\textit{DATACOMP: In search of the next generation of multimodal datasets}} by \citet{gadre2024datacomp} has completely different goals from Neural Priming, but achieves them similarly. The paper introduces DataComp, which is a test bed for dataset-related experiments that contains 12.8 billion image-text pairs retrieved from Common Crawl. Upon retrieving this pool, they proceed to train a new clip model with a fixed architecture and hyper-parameters. The paper concludes that CommonPool and LAION-2B are comparable with the same filtering. This means that image-based filtering and CLIP score filtering excels on most tasks, and can be effectively used to retrain other models. Despite this, the paper mentions that they found demographic biases in models trained using their pool, but their goal was not to reduce these harmful biases. In this paper we aim to offset this demographic bias found in models trained on large-scale filtered data pools such as DataComp.

\subsection{Ethics and Fairness}
\begin{figure}[H]
    \begin{center}
        \includegraphics[width=8cm]{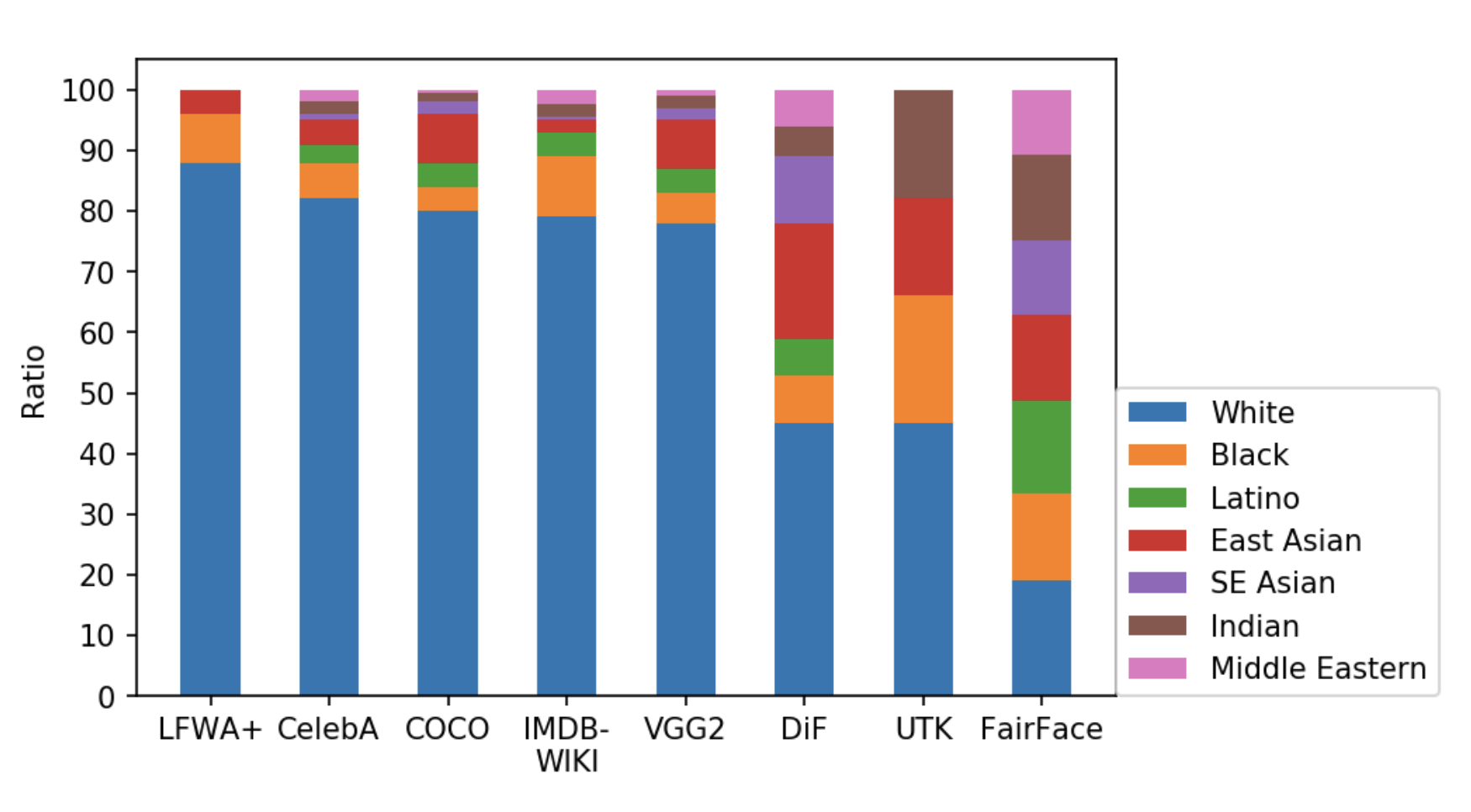}
        \caption{Racial compositions in face datasets \cite{karkkainen2021fairface}}
        \label{fig:face_datasets}
    \end{center}
\end{figure}

The FairFace dataset and classifier were first published in \textbf{\textit{FairFace: Face Attribute Dataset for Balanced Race, Gender, and Age for Bias Measurement and Mitigation}} by \citet{karkkainen2021fairface}. This project focused on creating a dataset and classifier that were balanced across race, gender, and age as shown in \cref{fig:face_datasets}. This balance is crucial because the paper demonstrates that the balance allows for improved generalization classification performance on the defined demographics, even on novel datasets that contain more non-White faces than typical datasets. The fact the simply balancing these demographics allows for increased accuracy and generalizability is extremely important. This is the core of D3G, and FairFace shows that balancing demographics results in performance improvements. The primary difference is that we aim to show similar improvements without any additional training. Alongside creating a balanced dataset, they also demonstrated their classifier produces balanced accuracy across the specified demographics, which is crucial because I use this classifier to create new labels for the IdenProf dataset. \\

\section{Methods}

\begin{figure*}
    \begin{center}
        \includegraphics[height=8cm]{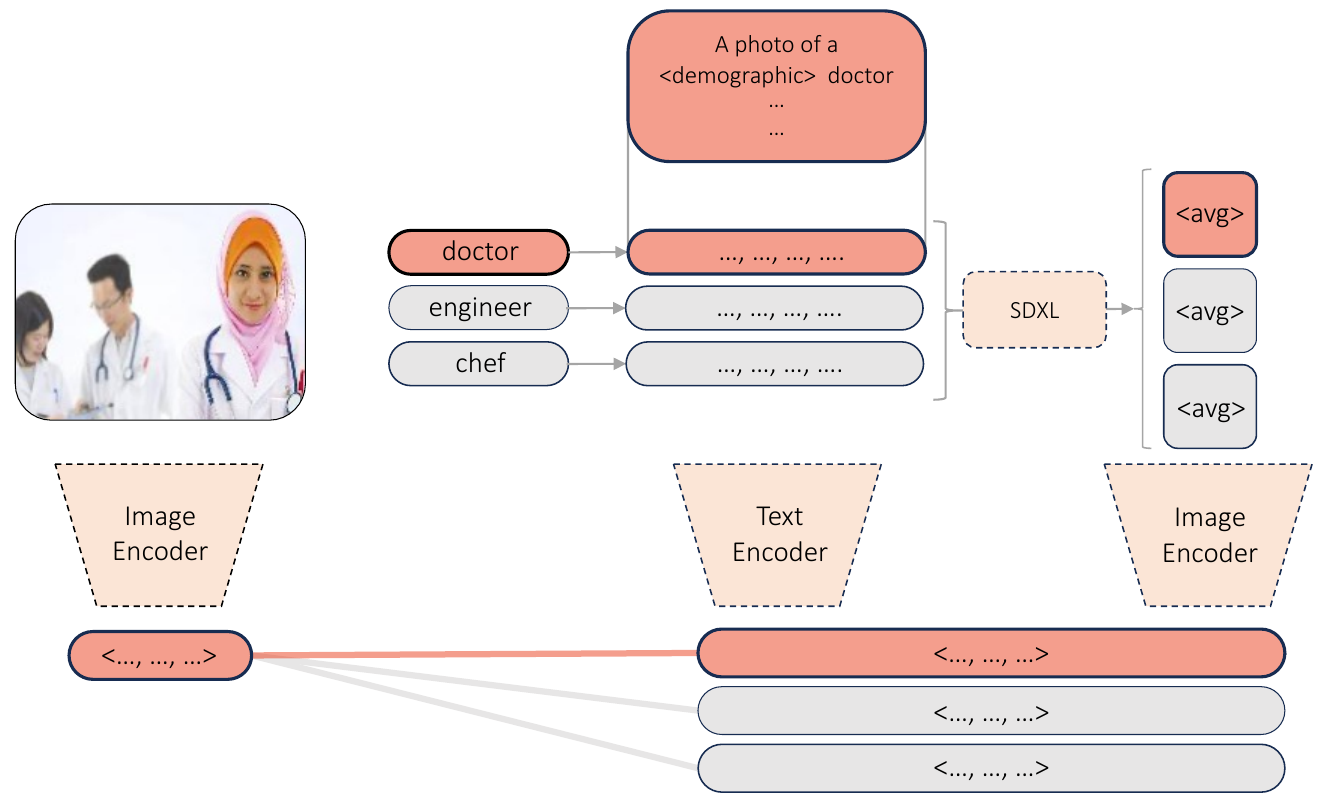}
        \caption{The D3G Framework}
        \label{fig:d3g}
    \end{center}
\end{figure*}

We aim to create an ensemble of models to improve multimodal image classification accuracy, especially for models that are trained on data with a class imbalance. We test this method on standard benchmark datasets, such as ImageNet as shown in \cref{fig:d3g_demo}, then we expand our technique to classify demographic-focused datasets. CLIP (Contrastive Language-Image Pretraining) \cite{radford2021learning} will be used for image-to-text retrieval, and Stable Diffusion XL 1.0 \cite{podell2023sdxl} for image generation. Our approach is as follows:

\subsection{Datasets}
For all the results shown in this paper, we classify images from the IdenProf test dataset. We selected this dataset because it provides a simple, applicable downstream task and because all the images were collected and filtered by hand via Google Image search. Each image in the dataset can belong to one of ten classes: Chef, Doctor, Engineer, Farmer, Firefighter, Judge, Mechanic, Pilot, Police, or Waiter. In total, there are 2,000 images for testing, with 200 images for each class. Finally, it is important to note the demographic distribution published by the dataset authors. The IdenProf dataset consists of 80.6\% male subjects, and 19.4\% female. Along with this, 91.1\% of the people within the dataset are White, while 8.9\% are of another race. The dataset author also notated that there were more images of Asian and White people obtainable, when compared to that of black people. Similarly, there were more images of men obtainable than of women. This reflects this demographic biases discussed previously.\\
\\
Along with IdenProf, we also leverage information collected from the FairFace dataset \citet{karkkainen2021fairface}. This dataset defines common demographics and forms them into classification categories. The authors constructed their dataset containing 108,501 images, and even though we do not utilize this dataset within this paper, the demographic information is still useful. We leverage the classification model that they trained on their own dataset. As a result, the results from the classifier are highly balanced and less likely to contain demographic bias. We use this classifier to assign additional labels to the images within IdenProf. There are 3 primary demographics that will be assigned as labels: race, gender, and age. Along with this, the race category has two versions, with race 4 being coarse-grained with only four races to choose from, and race 7 being fine-grained with 7 races to choose from. Combining the classes from IdenProf and FairFace, every image in the dataset can be classified with any of the labels identified in \cref{tab:d3g_classes}.
\\
\begin{table}
    \begin{center}
        \begin{tabular}{c|c}
            \textbf{Class} & \textbf{Values}\\
            \hline
            profession & Chef, Doctor, Engineer, Farmer, Firefighter,\\
             & Judge, Mechanic, Pilot, Police, or Waiter\\
            \hline
            race 7 & White, Black, Indian, East Asian, \\
                    & Southeast Asian, Middle Eastern, and Latino\\
            \hline
            race 4 & White, Black, Indian, Asian\\
            \hline
            gender & Male, Female\\
            \hline
            age & 0-2, 3-9, 10-19, 20-29, 30-39, \\
                & 40-49, 50-59, 60-69, 70+\\
        \end{tabular}
        \caption{All potential classes for an image from IdenProf}
        \label{tab:d3g_classes}
    \end{center}
\end{table}

\begin{table*}
    \begin{center}
        \begin{tabular}{|c|c|c|}
            \hline
            \textbf{Demographic} & \textbf{Prompt} & \textbf{Text}\\
            \hline
            Profession & "A photo of a $<$prof$>$" & A photo of a doctor\\
            \hline
            Race 7 & "A photo of a $<$race$>$ $<$prof$>$" & A photo of a white doctor\\
            \hline
            Race 4 & "A photo of a $<$race$>$ $<$prof$>$" & A photo of a white doctor\\
            \hline
            Gender & "A photo of a $<$gender$>$ $<$prof$>$" & A photo of a male doctor\\
            \hline
            Age & "A photo of a $<$age$>$ year old $<$prof$>$" & A photo of a 30-39 year old doctor\\
            \hline
        \end{tabular}
        \caption{Example diverse demographic texts for classifying \textbf{\textit{profession}}. Note that each prompt starts with "A photo of a," and that all the correct nouns and adjectives are added to the prompts as shown in the right column. More examples are provided in \cref{sec:appendix}.}
        \label{tab:d3g_prompts}
    \end{center}
\end{table*}

\subsection{Creating Prompts}
To generate our prompts, we leverage a set of templates constructed based on demographics identified in \cref{tab:d3g_classes}. These templates are designed to expose and leverage a specific demographic bias, based on the whatever image is currently being classified. For instance, if we were attempting to classify the profession of the person within the image, our prompts would be as shown in \cref{tab:d3g_prompts}. This process is pictured within \cref{fig:d3g}.

\subsection{Generate Class Images}
Upon creating diverse demographic prompts from templates for each of the classes, each of these prompts are used to generate an image. We employ Stable Diffusion XL, a diffusion-based image generation model, to conditionally generate an image of each class in the dataset. Our result will be images that emphasizes the diverse demographics between the classes. For standard D3G we will generate 1 image per prompt then average the embeddings of all the prompts, and for average image D3G we will generate 5 images per prompts, then perform the same process of averaging the embeddings of these images. Generating these diverse images is crucial because our goal is to combat the issue of prediction bias by generating diverse images and utilizing them for the next step when predicting labels. This step is depicted by images on the right of \cref{fig:d3g_demo}.

\subsection{Weighted Sum}
Using the prompts created earlier, we start the classification phase. We use the image and text encoders from CLIP ViT-L/14, our multimodal model, in order to get the embeddings for the generated images. Upon getting these embeddings, we scan values from 0 to 1 using a step value of 0.01 in order to find an optimal weight to create a weighted sum of the text and image embeddings. The text embedding will have a weight of $w$ while the image embedding is weighted by $1 - w$. This step allows us to bridge the semantic gap between text and images, because images are always closer in embedding space to one another than text. After performing this step, we will get a new embedding that represents the weighted combination of the text and image embeddings.

\subsection{Classification}
\begin{figure}[h!]
    \begin{center}
        \includegraphics[width=8.2cm]{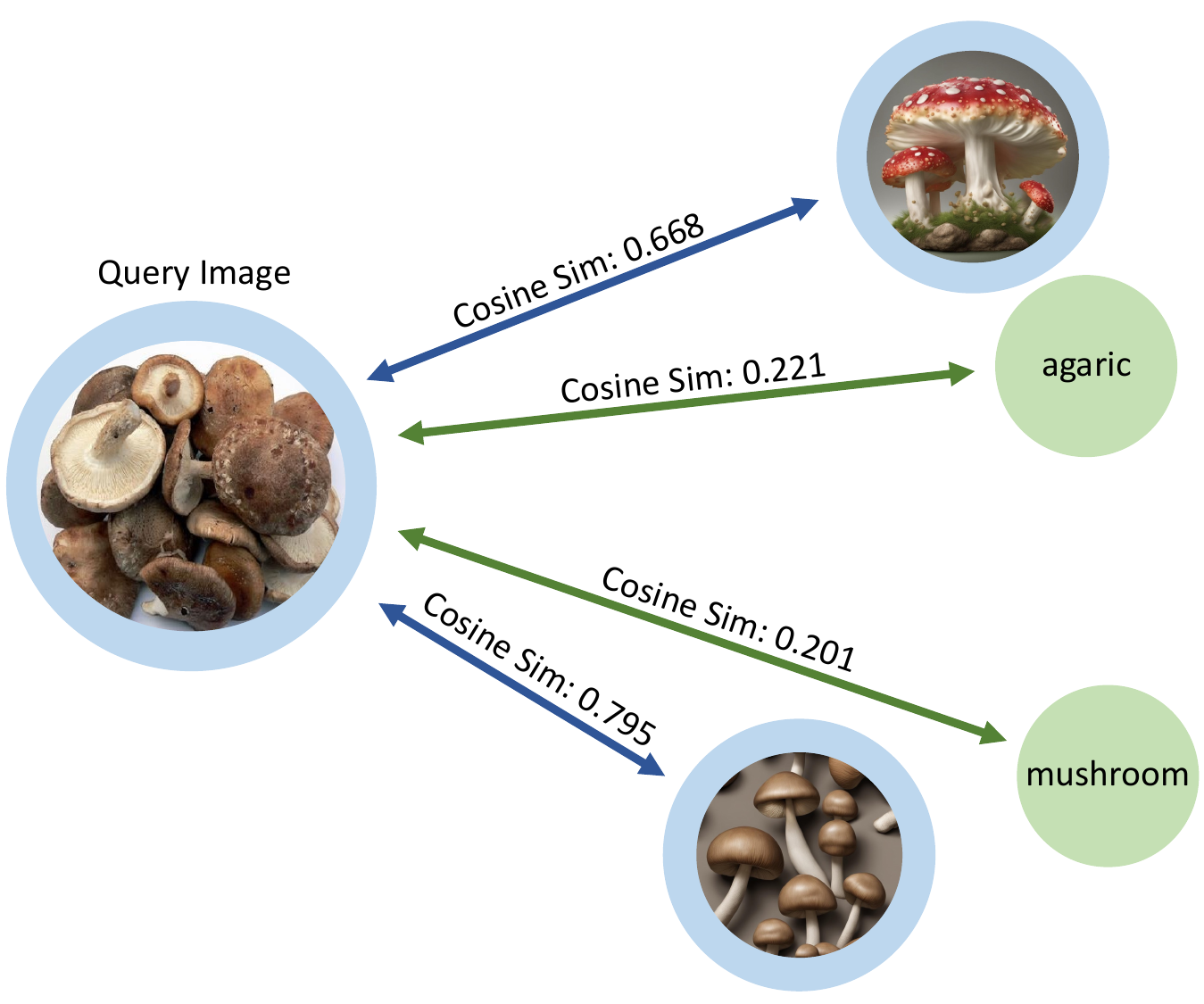}
        \caption{A demo example of D3G on difficult fine-grained classes from the ImageNet dataset (Note: we utilized ImageNet for this example to showcase the fine-grained classification capabilities of D3G. IdenProf does not have such fine-grained classes).}
        \label{fig:d3g_demo}
    \end{center}
\end{figure}

Finally, we get the embedding of the query image by passing it through the CLIP image encoder. At this point, we simply getting the cosine similarity between the query image embedding, and the combined image-text embeddings from each class. In order to classify the image, we just get the highest similarity score and use that class as the prediction. This step is depicted by the blue arrows within Figure \cref{fig:d3g_demo}.

\section{Results}

\subsection{Metrics}

We choose to use top-1 accuracy as the standard metric for our results. We selected this metric for a variety of reasons, the most prominent being that this paper aims to increase zero-shot classification accuracy. There are many metrics that represent zero-shot accuracy; however, top-1 accuracy is the most common.

\begin{figure}[h!]
    \begin{center}
        \includegraphics[width=8cm]{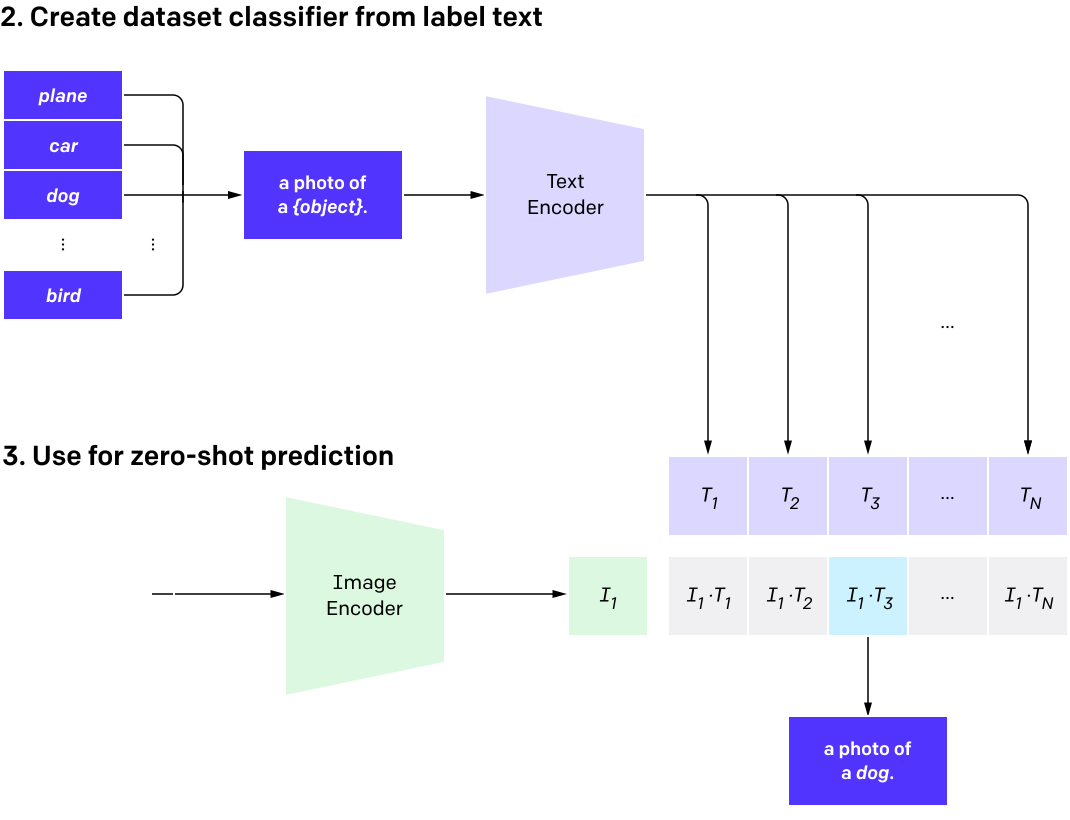}
        \caption{The process of classifying an image with CLIP at inference-time \cite{radford2021learning}.}
        \label{fig:clip}
    \end{center}
\end{figure}

\begin{table*}
    \begin{center}
        \begin{tabular}{|c|c|c|c|c|c|c|}
            \hline
             \textbf{Demographic} & \textbf{Method} & \textbf{Profession} & \textbf{Race 7} & \textbf{Race 4} & \textbf{Gender} & \textbf{Age}\\
            \hline
            \multirow{3}{*}{Profession} & CLIP & \cellcolor{gray!10} 95.14 & \cellcolor{gray!10} 94.73 & \cellcolor{gray!10} 95.22 & \cellcolor{gray!10} 96.52 & \cellcolor{gray!10} 94.81\\
            \cline{2-7}
            &Standard D3G & 95.54 & 95.22 & 95.30 & 96.52 & 95.06\\
            \cline{2-7}
            & Average Image D3G & \cellcolor{green!20} 95.87 & \cellcolor{green!20} 95.62 & \cellcolor{green!20} 95.38 & \cellcolor{green!20} 96.76 & \cellcolor{green!20} 95.54\\
            \Xhline{5\arrayrulewidth}
            \multirow{3}{*}{Race 7} & CLIP & \cellcolor{gray!10} 44.65 & \cellcolor{gray!10} 28.20 & \cellcolor{gray!10} - & \cellcolor{gray!10} 28.61 & \cellcolor{gray!10} 25.69\\
            \cline{2-7}
            &Standard D3G & 45.38 & 31.85 & - & 32.90 & 30.96\\
            \cline{2-7}
            & Average Image D3G & \cellcolor{green!20} 45.46 & \cellcolor{green!20} 32.33 & \cellcolor{green!20} - & \cellcolor{green!20} 33.55 & \cellcolor{green!20} 32.25\\
            \hline
        \end{tabular}
        \caption{Results when classifying the specified demographic of the people within the IdenProf dataset. The far left column shows the demographic that will be classified. The second column on the left dictates the method used for classification, while the other columns dictate the prompting structure, as discussed in \cref{sec:eval_breakdown}. Additional results are shown in \cref{sec:appendix}.}
        \label{tab:d3g_classify_prof}
    \end{center}
\end{table*}

\begin{table*}
    \begin{center}
        \begin{tabular}{|c|c|c|c|c|c|c|}
            \hline
             \textbf{Demographic} & \textbf{Method} & \textbf{Profession} & \textbf{Race 7} & \textbf{Race 4} & \textbf{Gender} & \textbf{Age}\\
            \hline
            \multirow{2}{*}{Profession} &Standard D3G & 0.85 / 0.15 & 0.84 / 0.16 & 0.90 / 0.10 & 0.91 / 0.09 & 0.90 / 0.10 \\
            \cline{2-7}
            & Average Image D3G & 0.71 / 0.29 & 0.74 / 0.26 & 0.84 / 0.16 & 0.91 / 0.09 & 0.67 / 0.33 \\
            \Xhline{5\arrayrulewidth}
            \multirow{2}{*}{Race 7} & Standard D3G & 0.90 / 0.10 & 0.68 / 0.32 & - & 0.68 / 0.32 & 0.67 / 0.33 \\
            \cline{2-7}
            & Average Image D3G & 0.92 / 0.08 & 0.69 / 0.31 & - & 0.67 / 0.33 & 0.68 / 0.32 \\
            \hline
        \end{tabular}
        \caption{The weight values used to achieve the results in \cref{tab:d3g_classify_prof}. For each evaluation, the left value is the text embedding weight, and the right is the image embedding weight. CLIP is not included because no images are weighted with the text embeddings. Note that the sum of the text and image weights for a given evaluation should equal 1.}
        \label{tab:d3g_weights}
    \end{center}
\end{table*}

\subsection{Evaluation Breakdown}
\label{sec:eval_breakdown}
We study three primary classification methods in this paper. CLIP ViT-L/14 is our baseline, the standard method of multimodal classification as outlined in \cref{fig:clip}. The second method of image classification is Standard D3G as shown in \cref{fig:d3g}. In this method, for every class in our dataset, we generate one prompt for each of the specified demographics, then use these prompts to generate images and average their embeddings. Finally, our third method is Average Image D3G. This is the exact same process as Standard D3G; however, instead of generating one image per demographic prompt, we will generate 5, then average the embeddings of all the prompts for a given class.\\
\\
Along with our three classification methods, we outline 5 prompting strategies when creating our demographic prompts within the D3G framework, as shown in the top row of \cref{tab:d3g_classify_prof}. It is important to note that for all of these prompting strategies, they \textbf{\textit{add}} demographic information, in addition to the specified classification category. For instance, as shown in \cref{tab:d3g_prompts}, when the task is to classify profession, we can add information regarding race or gender in addition to the standard profession class. This allows us to study how specific demographics affect the classification accuracy.\\
\\
Finally, we also explore and analyze the per-class accuracy results when classifying the specified demographic, as discussed in later sections.

\begin{table*}
    \begin{center}
        \begin{tabular}{c|c|c|c|c|c|c|c|c|}
            \hline
              \textbf{Demographic} & \textbf{Prompt} & \textbf{White} & \textbf{Black} & \textbf{Latino} & \textbf{East Asian} & \textbf{South East Asian} & \textbf{Indian} & \textbf{Middle Eastern}\\
            \hline
            \multirow{5}{*}{Race 7} & Profession & \cellcolor{green!20}68.19 & 70.90 & \cellcolor{green!20}15.38 & 43.46 & \cellcolor{green!20}20.59 & 57.58 & 13.80\\
            \cline{2-9}
            & Race 7 & 8.92 & 66.42 & 11.54 & \cellcolor{green!20}52.74 & 8.82 & 60.61 & 32.68\\
            \cline{2-9}
            & Race 4 & - & - & - & - & - & - & -\\
            \cline{2-9}
            & Gender & 18.07 & \cellcolor{green!20}73.13 & 11.54 & 35.02 & 17.65 & 51.52 & \cellcolor{green!20}34.93\\
            \cline{2-9}
            & Age & 12.29 & 68.66 & 11.54 & 43.46 & 14.71 & \cellcolor{green!20}69.70 & 29.58\\
            \hline
        \end{tabular}
        \caption{\textbf{\textit{Standard D3G}} per-class results when classifying the specified demographic. Note that all the prompts are as described in \cref{tab:d3g_prompts} (e.g. "A photo of a black person", or "A photo of a 30-39 year old doctor"). More examples are provided in \cref{sec:appendix}}
        \label{tab:d3g_race_per-class}
    \end{center}
\end{table*}

\subsection{Top-1 Results}
The top-1 results when classifying two out of the 5 demographics are shown in \cref{tab:d3g_classify_prof} (Additional results for the other demographics will be included in \cref{sec:appendix}).\\
\\
CLIP performs fairly well when classifying profession, which was to be expected because CLIP's training data likely includes richer cross-modal representations related to profession. As a result, all the accuracies are quite high. Despite this already high performance, D3G is able to still improve performance. This implies that providing diverse demographics can still improve CLIP's understanding of well-established concepts.\\
\\
We are able to gain a better understanding of D3G's efficacy when we look at the performance gains on Race 7. For this demographic, CLIP performance is much worse. Once again, the model performs better when the prompt contains information regarding profession, due to the increased likelihood of this information within the training data; however, when this information is omitted, the performance on the other demographics is abysmal. With the accuracy only scoring around 10-15\% more than random guessing (which should be around 14\% top-1 accuracy), this shows that CLIP does not have a deep understanding of race and other demographics.\\
\\
With this in mind, by simply implementing D3G, we are able to push the accuracies up to by 4-7\%. This indicates that coupling the standard embeddings with diverse data that has been generated, improves CLIP's understanding of concepts that were previously misunderstood. In addition, it is important to note that Average Image D3G typically performs better than the standard method. Once again, this makes sense and conforms to our hypothesis. Generating diverse data pushes the embeddings closer to the ground-truth position within embedding space, resulting in more accurate predictions for classes the model may not fully understand.\\
\\
These results are highly promising, and we can learn a bit more about the effect of D3G on these results by looking at the weights utilized to produce these scores.

\subsection{Weighting Strategy}

Recall that in order to classify a query image, we form a weighted sum of the embeddings between the text prompt, and the generated images. The ratio of text-to-image weighting is dictated by scanning values until an optimal state is found. This is necessary, because for certain images, the text embeddings will contribute more to the classification result than the image embeddings and vice versa. With this in mind, the weighting ration between images and text is also an indicator of how much the generated images from D3G actually help given a defined demographic. Knowing this information, we can then start to understand exactly what our results mean in the broader scope.\\
\\
When viewing the weights from \cref{tab:d3g_weights}, we see the same trends that were displayed within \cref{tab:d3g_classify_prof}, but we get a glimpse as to why D3G had minimal performance gains. When classifying profession, most of the text weights for Standard D3G are quite high, being roughly around 85-90\%; however, whenever see larger increases in accuracy for D3G, we also see an increased weighting of the generated images. This is especially evident when classifying race 7. Once again, the prompts that utilized professions were able to get somewhat higher accuracies, due to the structure of the dataset; however, for every other race 7 evaluation, the generated images played a major role in the classification results. The fact that images were consistently weighted around 30\% shows that the diversity matters when classifying demographics.\\
\\
The results analyzed from our top-1 results and their corresponding weighting strategies, show \textit{that} the method works; however, the per-class results give us a deeper understanding of \textit{why} the method works. 

\subsection{Per-Class Results}
For these results, we primarily reference \cref{tab:d3g_race_per-class}; however, note that additional per-class results are included within \cref{sec:appendix}. When classifying race 7, we know that the best performance gains were from including gender and age into the prompts. Focusing on these rows, we can see a few interesting trends. For instance, including information about gender improves the accuracy for black and middle eastern people the most. This is likely due to the fact that within CLIP's training data, these populations have gender underrepresented. Within \cref{sec:future_work}, we will later discuss future methods of confirming this hypothesis.\\
\\
Now that we understand which demographics help classification accuracies, we can now start to extend these inferences across demographics. Images and text related to East Asian people likely did not have rich cross-modal representations because race 7 helped the most for this demographic. This means that simply generating images of diverse races was able to significantly boost the accuracy. Similarly, age was the most useful demographic when classifying people in images as Indian. This was quite surprising, and as we discuss in \cref{sec:future_work}, we intend to further explore the impact of these results by including additional metrics such as precision, recall, specificity, and F1 score.\\
\\
Another trait of these per-class results emerges when we compare the accuracy ratios across demographic columns. For instance, black generally achieves a higher per-class accuracy than the other demographics, with Indian and East Asian obtaining the second and third-highest overall per-class accuracies across all the prompts. Alongside this, Latino, South East Asian, and White achieve some of the lowest per-class accuracies overall across all prompts. We were very surprised by this outcome, especially by the fact that race 7 was the worst performing prompt for White, which had the majority representation within the dataset. Intuitively, this may imply that providing diverse representations can also move embeddings away from the correct position in embedding space. In order to combat this, we may be able to strategically weight generated image and text prompt embeddings in relation to their demographic proportions within the dataset (e.g., if Latino is underrepresented within the dataset, then we will up-weight the Latino embeddings). This idea is further explored within \cref{sec:future_work}.\\
\\
Finally, we did not describe the results for profession, due to the fact that we cannot infer why these demographics performed best, due to the fact that CLIP leverages profession information to make its predictions, but the dataset is catered towards profession. This means that the increased accuracies could be either due to the profession information within the prompt, or the images generated of each profession. Either way, we will need to run more tests to fully understand this. We intend to evaluate on other datasets, so we can understand whether this correlation indicates causation; however, these are very promising results.

\section{Discussion}
\subsection{Assumptions}
Within this paper, two prominent assumptions are made:
\begin{enumerate}
  \item The generative model has a better learned representation of the true distribution of the data (due to its increased complexity and data diversity).
  \item The base multimodal model can distinguish between similar classes. Our method will not improve performance if this is not the case.
\end{enumerate}
These assumptions are necessary for D3G to function properly, but they are not unreasonable for a zero-shot setting. The generative model must have a better learned representation of the true data distribution, because it needs to be able to generate images that accurately represent the desired concept. If the model cannot generate useful images, then D3G will revert to using the baseline CLIP method, with text-based classification.\\
\\
In addition, we need our base model to be able to distinguish between similar classes, because if two classes correspond to the same point within embedding space, then our model cannot distinguish them. Similarly, we need this assumption so that the weighted sum of the image and text embeddings actually pushes the embedding towards the true embedding, and not just in a random direction. If the base model couldn't distinguish between certain classes, then we would have no guarantee that creating a weighted sum actually improves classification, because the model would be completely guessing in that case. In the future, we may be able to validate this assumption by comparing the embeddings within embedding space to ensure they are an adequate distance apart, but for now this will be maintained as an assumption.\\
\\
These assumptions on their own are not unreasonable; however, in certain circumstances they may become limitations, as discussed later.\\
\\
As mentioned previously, this research is crucial because models such as CLIP, dictate are frequently used to filter large datasets such as DataComp-1B or LAION-2B. If CLIP performs so poorly when classifying demographics, then these biases will be reinforced on all models trained with the datasets. This issue has compounding effects, and so to reduce demographic bias within image generation, object detection, and content moderation models, we \textbf{\textit{must}} start with image classification.

\subsection{Future Work}
\label{sec:future_work}
With such promising results from this project, there are many steps we intend to take in the future, in order to ensure this method is as robust as possible.\\
\\
To start, we aim to include additional metrics that properly quantify the balance between demographics to better understand how D3G balanced the predictions of the multimodal classifier. We specifically hope to investigate the robustness of our approach to class imbalance, data redundancy, and noise levels.\\
\\
For this paper, we decided to simply average the embeddings of all images generated with D3G, however this may not be the most effective process. Even though we generate images of a diverse range of demographics, these demographics are not weighted equally by CLIP (as demonstrated previously in \cref{tab:d3g_race_per-class}), due to the training data. This means that by utilizing the CLIP image encoder to get embeddings for all of our images, we are only offsetting the existing bias, but this does not create a neutral embedding; rather, it creates an embedding that still emphasizes the existing bias but is slightly more balanced across demographics. In order to combat this, we aim to explore how we can create a weighted sum of the embeddings from individual images, that is informed by the demographics of the training data and of the broader world. Intuitively, if CLIP tends to favor one demographic, then we will down weight those images, and vice versa if CLIP rarely selects another demographic. In this way, we can robustly enforce equity within CLIP's predictions.\\
\\
In addition to this step, in the future we also aim to utilize OpenCLIP so that we can accurately draw conclusions about the model's predictions in relation to the training data. Since we solely used CLIP as a baseline for this paper, we are unable to confidently state that the distribution of the training data led to the model's sometimes biased predictions; however, this is strongly implied. By utilizing a model with open training data and architecture, we can draw these conclusions with certainty. Researchers are starting to explore demographic bias within LAION-2B and DataComp-1B (the training data for certain OpenCLIP models), and we aim to leverage this knowledge for future implementations.\\
\\
We would also like to expand our evaluation suite to multiple datasets. Currently, we only evaluate on 2,000 images from IdenProf, but we could start by utilizing the full dataset of 11,000 training and test images, since we are not training, and we want a wider pool of images. In addition, we intend to perform similar tests over the FairFace dataset. These results would more effectively isolate CLIP's capabilities in predicting the demographics outlined in this paper, since the FairFace dataset was constructed with these demographics in mind. This is especially important, because we found that CLIP was able to leverage the semantic information regarding professions within the dataset, in order to classify race 7 more accurately. By removing professions as a factor, we will be able to fully explore CLIP's performance on such tasks.\\
\\
An important note, was that we were particularly intrigued by CLIP's inadequate performance when classifying demographics such as race 7, so we also aim to conduct an analysis on the individual classification results, combined with metrics such as precision, recall, specificity, and overall F1 score in order to better understand whether CLIP's performance on these demographics is statistically significant. If the positive predictions are solely informed by demographic stereotypes, then we aim to expose these weaknesses and combat them with D3G.\\
\\
Finally, in addition to generating images based on the demographics, we also aim to explore methods of retrieving images, or modifying the demographics of the query image in-place. Modifying the existing query image to get diverse demographics, may reduce the impact of stereotypes enforced by the image generation model, and result in classifications that are much more accurate.

\section{Conclusion}
Image classification remains a challenging task despite advancements in multimodal models like CLIP that leverage semantic similarities across vision and language. Low-capacity models often suffer from underfitting, leading to poor performance; however, the generation of high-quality data with rich cross-modal representations is also difficult. Imbalanced demographics in datasets can cause predictions to bias toward more represented classes, pushing those who are underrepresented to the wayside. Our study highlights these issues and their impact on zero-shot image classification, proposing Diverse Demographic Data Generation (D3G) as a solution. This training-free, zero-shot method enhances classification accuracy and reduces demographic bias in pre-trained multimodal models by providing diverse demographic data at inference time, demonstrating improved performance for these models.

\section{Ethics Statement}
The fact that we are utilizing image generation models for D3G provides significant potential for negative societal impact. For instance, the images generated by the model can often reinforce certain demographic biases. This is to be expected, because the prompts used within this paper are quite vague; however, this also shows that the generative model has learned visual stereotypes from its training data. The stereotypes within the generated images is why they should only be used as a weighted sum with the text, and never as the sole ground-truth signal. Excess up-weighting of the images, provides opportunity for unethical image generations.\\
\\
One potential way to combat this issue of stereotypes within generation, is to utilize the method discussed in \cref{sec:future_work}, where we modify the query image in-place in order to reduce the room for error, while still increasing demographic diversity.\\
\\
Along with this, our use of generative modelling allows for potentially unethical prompting. The only restrictions on prompting are those enforced by Stable Diffusion XL; however, due to the open-source nature of the model, many of these restrictions can be circumvented. We do not condone the use of D3G to generate any hateful, demeaning, or otherwise unethical data. This method should only be used within appropriate contexts, and primarily as a means of increasing pre-trained model diversity ad hoc.\\
\\
The selection of demographics used within our classification process was mainly a result of the process used to create the FairFace dataset \cite{karkkainen2021fairface}. The authors defined the races used to be based on commonly accepted race classification from the U.S. Census Bureau; however, we acknowledge that does not properly represent the racial landscape of the world. It is important to note that the authors decided to use skin color as a proxy to race, combined with annotations about physical attributes. This means that the annotations used to construct the dataset and train the FairFace classification model used to create labels for IdenProf, may contain annotator bias. This is evident in the gender demographic. The authors mentioned it would be impossible to perfectly balance the gender predictions of their model, outside a lab setting. Finally, ages were simply segmented into common age groups. The decision to use these demographic categories limits the conclusions we can draw in this paper, regarding the impact of all relevant demographics on classification accuracy.\\
\\
Finally, D3G is a technique that \textbf{\textit{does not}} remove demographic biases, but rather, it offsets learned biases. This means that the method can either reduce or accentuate human bias, and should not be used as a universal architecture to improve multimodal model fairness and accuracy. If the images generated contain harmful bias, then this technique could make the performance worse and much more inequitable.

\section{Limitations}
Due to this paper being focused on classification, a significant limitation is with regard to demographic intersectionality. People that fit into multiple demographics within the same category (i.e., people who are biracial), will suffer from only being classified as a single demographic. This is a limitation, because it is a known issue that cannot be surmounted using standard metrics within image classification. Future methods may be able to explore intersectionality by retrieving the top-k classified demographics; however, this would be difficult in a zero-shot setting, where no additional information about the query image is provided.\\
\\
A second major limitation is the fact the D3G can only perform well if the pretrained models are able to effectively distinguish between the demographics being classified. As mentioned within previously, if the multimodal model embeds two demographics to the same point in embedding space, or if the image generation model cannot generate good images for a given demographic, the technique will fail. This is typically not an issue for the broad demographics covered within this paper; however, it may become more difficult as the classes become more fine-grained.\\
\\
A final limitation is the fact that D3G utilizes pre-trained models for every step of the pipeline. This partially is also the most useful part of the technique; however, it also means that the limitations of the pretrained models will extend to D3G. The abilities or inabilities of the generative model will result in the final classification accuracies. Similarly, the quality of the embeddings produced from the multimodal model will dictate the effect D3G will have on classification accuracy.

\clearpage 
\bibliography{main}
\clearpage 
\begin{center}
\section{Appendix}
\end{center}
\label{sec:appendix}
\begin{table*}[H]
    \begin{center}
        \begin{tabular}{|c|c|c|}
            \hline
            \textbf{Demographic} & \textbf{Prompt} & \textbf{Text}\\
            \hline
            Race 7 & "A photo of a $<$race$>$ person" & A photo of a black person\\
            \hline
            Race 4 & "A photo of a $<$race$>$ person" & A photo of a black person\\
            \hline
            Profession & "A photo of a $<$race$>$ $<$prof$>$" & A photo of a black doctor\\
            \hline
            Gender & "A photo of a $<$race$>$ $<$gender$>$" & A photo of a black male\\
            \hline
            Age & "A photo of a $<$age$>$ year old $<$race$>$ person" & A photo of a 30-39 year old black person\\
            \hline
        \end{tabular}
        \caption{Example diverse demographic texts for classifying \textbf{\textit{race 7}}. The texts for classifying race 4 will be identical, just fewer racial classes.}
        \label{tab:d3g_race_prompts}
    \end{center}
\end{table*}
\begin{table*}
    \begin{center}
        \begin{tabular}{|c|c|c|}
            \hline
            \textbf{Demographic} & \textbf{Prompt} & \textbf{Text}\\
            \hline
            Gender & "A photo of a $<$gender$>$" & A photo of a female\\
            \hline
            Profession & "A photo of a $<$gender$>$ doctor" & A photo of a female doctor\\
            \hline
            Race 7 & "A photo of a $<$race$>$ $<$gender$>$" & A photo of a black female\\
            \hline
            Race 4 & "A photo of a $<$race$>$ $<$gender$>$" & A photo of a black female\\
            \hline
            Age & "A photo of a $<$age$>$ year old $<$gender$>$ person" & A photo of a 30-39 year old female\\
            \hline
        \end{tabular}
        \caption{Example diverse demographic texts for classifying \textbf{\textit{gender}}.}
        \label{tab:d3g_gender_prompts}
    \end{center}
\end{table*}
\begin{table*}
    \begin{center}
        \begin{tabular}{|c|c|c|}
            \hline
            \textbf{Demographic} & \textbf{Prompt} & \textbf{Text}\\
            \hline
            Age & "A photo of a $<$age$>$ year old" & A photo of a 30-39 year old\\
            \hline
            Profession & "A photo of a $<$age$>$ year old doctor" & A photo of a 30-39 year old doctor\\
            \hline
            Race 7 & "A photo of a $<$age$>$ year old $<$race$>$ person" & A photo of a 30-39 year old black person\\
            \hline
            Race 4 & "A photo of a $<$age$>$ $<$race$>$" & A photo of a 30-39 year old black person\\
            \hline
            Gender & "A photo of a $<$age$>$ $<$gender$>$" & A photo of a 30-39 year old female\\
            \hline
        \end{tabular}
        \caption{Example diverse demographic texts for classifying \textbf{\textit{gender}}.}
        \label{tab:d3g_age_prompts}
    \end{center}
\end{table*}
\begin{table*}
    \begin{center}
        \begin{tabular}{c|c|c|c|c|c|c|c|c|}
            \hline
              \textbf{Demographic} & \textbf{Prompt} & \textbf{White} & \textbf{Black} & \textbf{Latino} & \textbf{East Asian} & \textbf{South East Asian} & \textbf{Indian} & \textbf{Middle Eastern}\\
            \hline
            \multirow{5}{*}{Race 7} & Profession & 68.92 & 70.90 & 15.38 & 43.46 & 20.59 & 57.58 & 13.24\\
            \cline{2-9}
            & Race 7 & 9.40 & 61.94 & 15.38 & 66.24 & 5.88 & 60.61 & 26.48\\
            \cline{2-9}
            & Race 4 & - & - & - & - & - & - & -\\
            \cline{2-9}
            & Gender & 20.72 & 67.91 & 15.38 & 44.73 & 11.76 & 51.52 & 29.86\\
            \cline{2-9}
            & Age & 11.57 & 65.67 & 11.54 & 59.92 & 14.71 & 69.70 & 25.07\\
            \hline
        \end{tabular}
        \caption{\textbf{\textit{Average D3G}} per-class results when classifying the specified demographic. Note that all the prompts are as described in \cref{tab:d3g_prompts}.}
        \label{tab:d3g_avg_race_per-class}
    \end{center}
\end{table*}
\begin{table*}
    \begin{center}
        \begin{tabular}{c|c|c|c|c|c|c|c|c|c|c|c|}
            \hline
              & Prompt & \textbf{Chef} & \textbf{Doctor} & \textbf{Engineer} & \textbf{Farmer} & \textbf{Firefighter} & \textbf{Judge} & \textbf{Mechanic} & \textbf{Pilot} & \textbf{Police} & \textbf{Waiter}\\
            \hline
            \multirow{5}{*}{\rotatebox{90}{Profession}} & Profession & 95.54 & 98.80 & 97.62 & 98.36 & 94.12 & 98.20 & 86.49 & 99.37 & 100.0 & 84.34\\
            \cline{2-12}
            & Race 7 & 94.27 & 98.80 & 96.43 & 98.36 & 96.08 & 99.40 & 81.08 & 98.74 & 100.0 & 84.94\\
            \cline{2-12}
            & Race 4 & 94.90 & 98.80 & 96.43 & 98.36 & 96.08 & 99.40 & 82.43 & 98.74 & 100.0 & 84.34\\
            \cline{2-12}
            & Gender & 96.82 & 98.80 & 96.43 & 96.72 & 98.04 & 99.40 & 90.54 & 99.37 & 100.0 &  87.35\\
            \cline{2-12}
            & Age & 97.45 & 98.19 & 96.43 & 98.36 & 98.04 & 98.80 & 85.14 & 97.48 & 100.0 & 80.72\\
            \hline
        \end{tabular}
        \caption{\textbf{\textit{Standard D3G}} per-class results when classifying the specified demographic. Note that all the prompts are as described in \cref{tab:d3g_prompts}. These results for race 7 are shown in \cref{tab:d3g_race_per-class}.}
        \label{tab:d3g_prof_per-class}
    \end{center}
\end{table*}
\begin{table*}
    \begin{center}
        \begin{tabular}{c|c|c|c|c|c|c|c|c|c|c|c|}
            \hline
              & Prompt & \textbf{Chef} & \textbf{Doctor} & \textbf{Engineer} & \textbf{Farmer} & \textbf{Firefighter} & \textbf{Judge} & \textbf{Mechanic} & \textbf{Pilot} & \textbf{Police} & \textbf{Waiter}\\
            \hline
            \multirow{5}{*}{\rotatebox{90}{Profession}} & Profession & 93.63 & 98.80 & 98.81 & 100.0 & 94.12 & 96.41 & 83.78 & 99.37 & 100.0 & 90.36\\
            \cline{2-12}
            & Race 7 & 93.63 & 98.80 & 97.62 & 98.36 & 94.12 & 97.60 & 81.08 & 98.74 & 100.0 & 90.36\\
            \cline{2-12}
            & Race 4 & 93.63 & 98.80 & 96.43 & 98.36 & 94.11 & 98.80 & 82.43 & 98.74 & 100.0 & 87.35\\
            \cline{2-12}
            & Gender & 96.18 & 98.80 & 96.43 & 98.36 & 98.04 & 99.40 & 90.54 & 99.37 & 100.0 & 89.16\\
            \cline{2-12}
            & Age & 94.90 & 98.80 & 95.24 & 100.0 & 94.12 & 96.41 & 82.43 & 95.60 & 100.0 & 92.77\\
            \hline
        \end{tabular}
        \caption{\textbf{\textit{Average D3G}} per-class results when classifying the specified demographic. Note that all the prompts are as described in \cref{tab:d3g_prompts}}
        \label{tab:d3g_avg_prof_per-class}
    \end{center}
\end{table*}

\end{document}